\documentclass{article}

\usepackage[preprint,nonatbib]{neurips_2023}

\usepackage[utf8]{inputenc} 
\usepackage[T1]{fontenc}    
\usepackage{hyperref}       
\usepackage{url}            
\usepackage{booktabs}       
\usepackage{amsfonts}       
\usepackage{nicefrac}       
\usepackage{microtype}      
\usepackage{xcolor}         

\usepackage{subfigure}

\usepackage{amsmath}
\usepackage{amssymb}
\usepackage{mathtools}
\usepackage{amsthm}
\usepackage{graphicx}
\usepackage{adjustbox}
\usepackage[capitalize,noabbrev]{cleveref}

\usepackage{graphbox}  
\usepackage{bm}
\usepackage{balance}
\usepackage{multirow}

\usepackage{wrapfig,lipsum,booktabs}

\DeclareMathOperator*{\argmax}{arg\,max}
\DeclareMathOperator*{\argmin}{arg\,min}
\newcommand{\best}[1]{\color{red}{\textbf{#1}}}
\newcommand{\secondbest}[1]{\color{blue}{#1}}

\theoremstyle{plain}

\theoremstyle{definition}

\theoremstyle{remark}

\usepackage{pifont}
\newcommand{\cmark}{\ding{51}}%
\newcommand{\xmark}{\ding{55}}%

\usepackage[textsize=tiny]{todonotes}
\newcommand{\mlnas}{\textsc{MP-NAS }}
\def\showcomments{0}
\newcommand{\hrushi}[1]{\if\showcomments1\textcolor{red}{[ HL: #1 ]}\else\fi}
\newcommand{\lukasz}[1]{\if\showcomments1\textcolor{blue}{[ LD: #1 ]}\else\fi}
\newcommand{\hongkai}[1]{\if\showcomments1\textcolor{yellow}{[ HW: #1 ]}\else\fi}
\newcommand{\abhi}[1]{\if\showcomments1\textcolor{red}{[ AB: #1 ]}\else\fi}

\title{How Much Is Hidden in the NAS Benchmarks? Few-Shot Adaptation of a NAS Predictor}

\author{
  Hrushikesh Loya\thanks{Indicates equal contributions} \\
  University of Oxford \\
  \texttt{loya@stats.ox.ac.uk} \\
  \And
  {\L}ukasz Dudziak$^{*}$ \\
  Samsung AI Center Cambridge \\
  \texttt{l.dudziak@samsung.com} \\
  \And
  Abhinav Mehrotra$^{*}$ \\
  Samsung AI Center Cambridge \\
  \texttt{abhinav.m1@samsung.com} \\
  \And
  Royson Lee \\
  Samsung AI Center Cambridge\\
  University of Cambridge \\
  \texttt{royson.lee@samsung.com} \\
  \And
  Javier Fernandez-Marques \\
  Flower Labs \\
  \texttt{jafermarq@gmail.com} \\
  \And
  Nicholas D. Lane \\
  Flower Labs \\
  University of Cambridge \\
  \texttt{ndl32@cam.ac.uk}
  \And
  Hongkai Wen \\
  Samsung AI Center Cambridge\\
  University of Warwick \\
  \texttt{hongkai.wen@warwick.ac.uk}
}

\begin{document}

\maketitle

\begin{abstract}
Neural architecture search has proven to be a powerful approach to designing and refining neural networks, often boosting their performance and efficiency over manually-designed variations, but comes with computational overhead.
While there has been a considerable amount of research focused on lowering the cost of NAS for mainstream tasks, such as image classification, a lot of those improvements stem from the fact that those tasks are well-studied in the broader context.
Consequently, applicability of NAS to emerging and under-represented domains is still associated with a relatively high cost and/or uncertainty about the achievable gains.
To address this issue, we turn our focus towards the recent growth of publicly available NAS benchmarks in an attempt to extract general NAS knowledge, transferable across different tasks and search spaces.
We borrow from the rich field of meta-learning for few-shot adaptation and carefully study applicability of those methods to NAS, with a special focus on the relationship between task-level correlation (domain shift) and predictor transferability; which we deem critical for improving NAS on diverse tasks.
In our experiments, we use 6 NAS benchmarks in conjunction, spanning in total 16 NAS settings -- our meta-learning approach not only shows superior (or matching) performance in the cross-validation experiments but also successful extrapolation to a new search space and tasks.
\end{abstract}

\section{Introduction}

Deep learning and neural networks have shown an exponential increase in applications ranging from computer vision, and natural language processing to more recently computational biology and climate sciences. Along with the wide success, there is often a tedious amount of manual ``artwork'' to design neural network architectures. Neural Architecture Search (NAS) tries to solve this problem by cleverly designing search spaces and strategies to output optimal neural architectures for a given task. NAS has already been applied to mainstream computer vision datasets like CIFAR10, CIFAR100, and ImageNet and is able to surpass hand-crafted neural networks with minimal computational overhead. Although its success on computer vision datasets, it was recently seen, most NAS algorithms didn't work well on other tasks, especially from other diverse domains \cite{tu2021bench}.

Recent works like XD operations \cite{roberts2021rethinking} and DASH \cite{shen2022efficient} have been proposed which expand the search space significantly by using convolution theorem to search for possible combination of kernel size and dilations in CNNs. These methods have been a first step towards building NAS methods for diverse tasks but do so by simply expanding the search space to look for an optimal architecture rather than improving on finding a neural architecture from a given search space. Furthermore, most NAS methods including XD operations and DASH do not benefit from the availability of large NAS benchmarks when applied to a new task. This means for a new task the NAS algorithm needs to train a large number of proposed architectures from scratch making it much slower for a diverse task.

We propose a novel NAS algorithm called \mlnas (short for meta-predictor-NAS) based on a Graph Convolutional Network (GCN) predictor that uses meta-learning to quickly adapt to new tasks and search spaces. Our approach observes competitive performance on the diverse tasks from NAS-Bench-360 \cite{tu2021bench} while also utilizing past neural network training from various NAS benchmarks (different search spaces and/or data modalities) for a much faster NAS algorithm. Specifically, we propose a predictor-based NAS framework where we use model-agnostic meta-learning \cite{finn2017model} to meta-learn the GCN predictor to quickly adapt and transfer its knowledge to rank architectures across different data modalities and search spaces.

We evaluate and compare our predictor's transfer ability on various search spaces, namely within TransNAS-Bench-101 (TB101) \cite{duan2021transnas}, NAS-Bench-201 (NB201) \cite{dong2020bench} and cross-search space from NB201 to TB101, NAS-Bench-101 (NB101) \cite{ying2019bench}, and NAS-Bench-ASR (NB-ASR) \cite{mehrotra2020bench}. We observe our predictor is able to provide more transferable predictions allowing for up to 10x speedup depending on the correlation between datasets and search spaces. As far as we know, this is also the first time cross-search space transferability (i.e. training and testing on two different search spaces) for NAS has been established. Finally, we apply our meta-learnt predictor on five NAS-Bench-360 (NB360) datasets along with a unified representation of the search space containing operations from previous cell-based NAS benchmarks to get substantial improvement compared to previous baseline NAS methods. We summarize our main contributions as follows:
\begin{itemize}
    \item We present the first attempt to leverage previous NAS results spanning different search spaces and tasks in a single process, to enable fast adaptation of the NAS predictor.
    \item As part of the above, we study applicability and effectiveness of the existing model-agnostic meta-learning approaches to predictor-based NAS.
    \item We present a way of unifying NAS search spaces, in order to allow using information from multiple NAS settings (exemplified in our work as NAS benchmarks).
    \item Despite simplicity of our setup in the aspects of search space design and training methodology, we show surprisingly strong performance on the challenging NAS-Bench-360 benchmark, showcasing the benefits of exploring better ways of utilising existing information.
\end{itemize}

\section{Related Work}
\textbf{Neural Architecture Search: }
NAS overcomes the problem of manually designing a neural network architecture by automating the process to output an optimal architecture for a given task. NAS approaches can be broadly classified into reinforcement learning based methods \cite{zoph2016neural, zoph2018learning, tan2019mnasnet}, evolutionary algorithms \cite{real2019regularized,guo2020spos,li2020dna} and gradient based methods \cite{liu2018darts,dong2019gdas,cai2019proxylessnas,li2020sgas,wang2021darts-pt}.
Furthermore, many recent NAS methods, irrespective of their search strategy, focus on improving efficiency by utilising performance predictors of various sorts to replace the time-consuming training.
These can again be broadly classified into methods using Bayesian optimisation \cite{shi2020bonas,white2020bananas}, graph neural networks (GNNs) \cite{wei2019neural_predictor,dudziak2020brpnas,wu2021weak_predictor} and, most recently, zero-shot predictors \cite{abdelfattah2021zero-cost,mellor2021nwot,ming2021zennas,chen2021neural}.
Along a similar line of work to ours, some authors focus on improving performance predictors by utilising, among other things, unlabeled data \cite{luo2020seminas,tang2020semi_assessor,liu2020labels,li2021generic}, predictor ensembles \cite{wu2021weak_predictor,zhao2023cold_start} or supernet pretraining \cite{cai2020onceforall,chen2021ctnas}.
We refer in more detail to relevant parts of prior work as needed throughout our paper.


\textbf{Neural Architecture Search for Diverse Tasks: }
NAS has mainly focused on speeding-up the search process on a few mainstream vision tasks, with generally a gap in performance of various NAS algorithms outside the mainstream vision tasks \cite{tu2021bench, colin2022adeeperlook}. NAS-Bench-360 was recently introduced as a benchmark-suite to bridge this gap and test for applicability of state-of-the-art NAS algorithms on ten diverse tasks from various domains including computer vision, genetics, geography, physics and health data. NAS-Bench-360 also showed the poor performance of some popular NAS methods like DARTS \cite{liu2018darts}, DenseNAS \cite{fang2020densely} and TE-NAS \cite{chen2021neural} when applied to diverse domains.

Recently, DASH \cite{shen2022efficient} and XD operations \cite{roberts2021rethinking} have been proposed to tackle this problem by leveraging Fourier convolutions and efficiently search an expansive search space.
This is done by either relaxing the discrete Fourier transform matrices by Kaleidoscope matrices in XD-operations or using a set of discrete convolution matrices calculated using fast Fourier transform. Both these methods try to solve the problem of NAS on diverse tasks by mainly expanding the search space to include larger convolution kernels, whereas we try to improve performance-predictor in NAS such that it can find the optimal architecture in any search space.
Furthermore, DASH and XD-operations ignore previous NAS training knowledge while performing NAS.
We believe and empirically show, accounting for previous NAS training from NAS benchmarks improves the performance of NAS significantly.



\textbf{Meta-learning: }
Meta-learning aims to train a neural network over a set of tasks such that it can quickly adapt to a new task. One of the most commonly used meta-learning algorithms include model-agnostic meta-learning (also known as MAML) \cite{finn2017model}, which adapts the model for a given task in the inner loop while simultaneously driving the model for a better initialization in the outer loop. There are many variants of the original MAML algorithm, for example Body Only Meta-Learning (BOIL) \cite{oh2020boil} which updates everything except the final layer in the inner loop, to enforce better feature representation.

\textbf{Meta-learning meets Neural Architecture Search: }
This includes, MetaNAS \cite{elsken2020meta} which uses DARTS \cite{liu2018darts} along with first-order meta-learning algorithm REPTILE \cite{nichol2018reptile} to not only adapt weights but also architectural parameters for improved few-shot learning. T-NAS \cite{lian2019towards} and BASE \cite{shaw2019meta} use MAML++ \cite{finn2017model, antoniou2018train} and Bayesian meta-learning similarly to MetaNAS to quickly adapt given the dataset to output architectures and weights.
MetaD2A \cite{lee2021rapid} is another meta-learning based NAS method recently proposed that uses amortized meta-learning along with set-encoder based graph generator which is jointly trained to output optimal neural architecture given a dataset.
On the other hand, HELP \cite{lee2021help} uses a similar amortized meta-learning framework to MetaD2A but in order to allow efficient few-shot adaptation of a latency predictor to new hardware devices.
Different hardware devices are represented with embedding vectors obtained from benchmarking a set of reference models, making the method not only focused specifically on latency estimation but also limited to a predefined search space.
All these methods leverage training information from other tasks but are not built for transferability to diverse tasks and are only tested in very similar datasets like CIFAR100, SVHN and ImageNet.
Additionally, similar to HELP but unlike other methods, we don't metalearn the architecture directly, rather the performance predictor - making our method easily applicable to different search strategies and spaces.
\section{Method}
\begin{figure*}
    \centering
    \includegraphics[width=\textwidth]{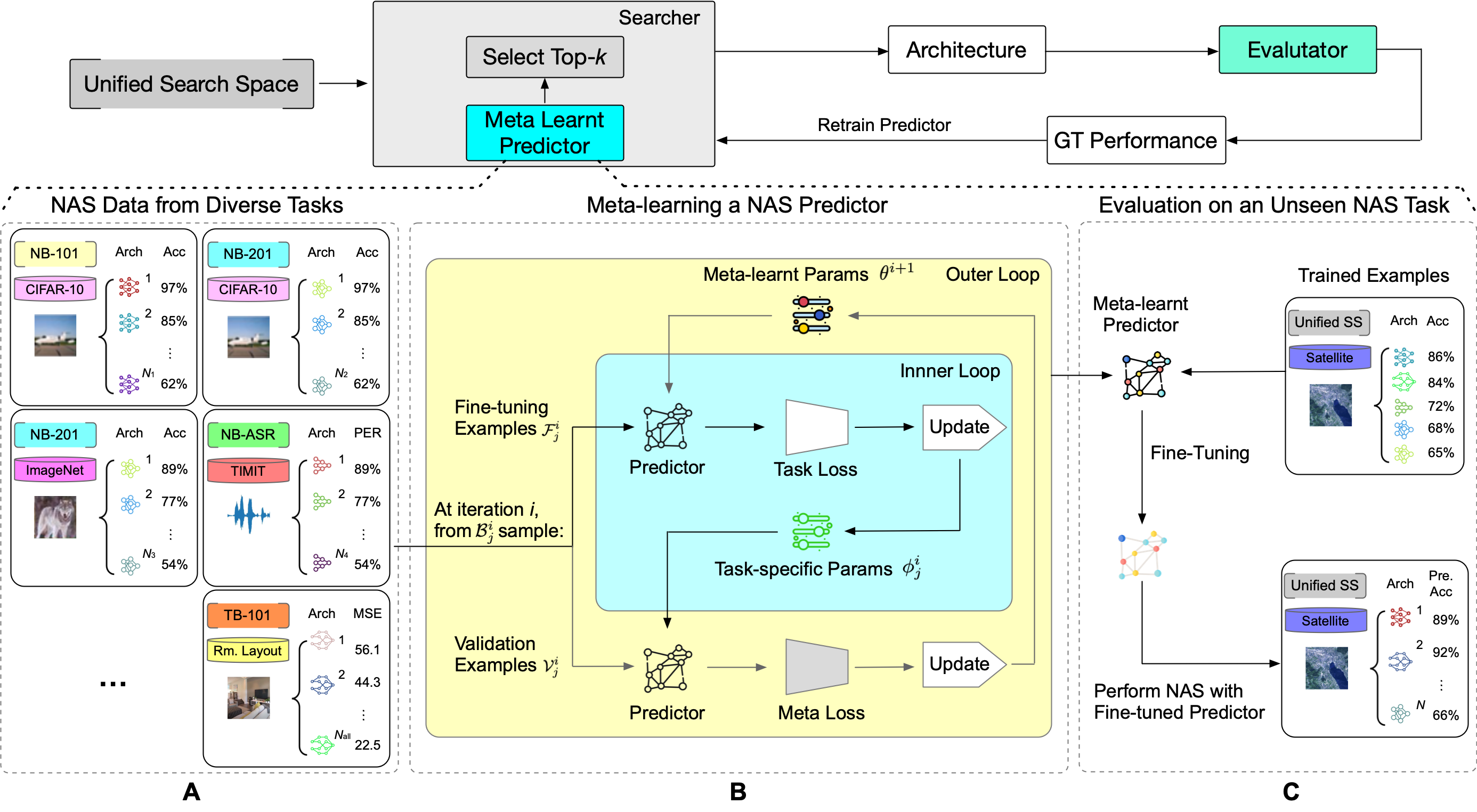}
    \caption{Overview of our method. \textbf{(A)} Combining previous NAS benchmarks and proposing a unified search space. \textbf{(B)} Meta initializing the GCN predictor for quick task adaptation. \textbf{(C)} Evaluating meta-learnt predictor on an unseen (new) NAS task.}
    \label{fig:overview}
\end{figure*}

We are interested to make a NAS algorithm that leverages the vast amount of previous NAS training to provide a more accurate performance predictor on a new dataset or even a new search space. We start by defining some notations which will become handy as we formalize our method. A overview of the method can be found in Figure \ref{fig:overview}.

\subsection{Preliminaries}

\noindent \textbf{NAS Tasks.}
Given a search space $\mathcal{A}$ and a dataset $D$, a NAS task $\mathcal{T}=\{\mathcal{A}, D, s\}$ is to discover the optimal neural architecture $A^{*}\in \mathcal{A}$ under a given performance metric $s$ when trained on $D$: $A^{*} = \argmax_{A \in \mathcal{A}} s(A; D)$, e.g., finding the neural network from NB201 search space that can offer the best object classification accuracy on CIFAR-10 dataset. Without loss of generality, in this paper we assume a neural architecture $A\in\mathcal{A}$ can be represented as a directed acyclic graph.

\noindent \textbf{NAS Data.}
For a given NAS task $\mathcal{T}=\{\mathcal{A}, D, s\}$, after training $N$ architectures on the dataset $D$, we can obtain the corresponding architecture-performance pairs $\{ A_i, s(A_i;D) \}_{i=0}^N$, which can be used in various ways by NAS algorithms, for example to train a NAS predictor.
We will use $\mathcal{B}_{\mathcal{T}}$ to refer to such a set for the NAS task $\mathcal{T}$.
Note that for readability, unless stated otherwise, henceforth we will use $s$ to refer to values of the performance metric (that is, $s(A;D)$) rather than the metric itself.

\noindent \textbf{NAS Predictors.}
A crucial component of most NAS methods concerns the evaluation of architecture performance (e.g., accuracy), which can be helpful in guiding the search, or used directly to perform the search.
For the latter, a popular approach is to train a performance predictor using the available NAS data, which is then used to predict the performance of new/unseen architectures.
Typically, a NAS predictor parameterized by $\theta$, denoted as $f_{\theta}$, is trained on the NAS data of a single NAS task $\mathcal{T}$, with the objective to minimize the loss between true and predicted performance:
\begin{equation}\label{eq:predictor_objective}
    \theta^{*} = \argmin_{\theta} \mathbb{E}_{(A,s) \sim \mathcal{B}_{\mathcal{T}}} \mathcal{L}(f_{\theta}(A); s)
\end{equation}
where $\mathcal{L}$ is the MSE loss between true and predicted performance, often minimized via gradient descent. Additionally, we consider a GCN-based performance predictor in our study due to its better performance on previous NAS tasks \cite{white2021performance_predictors} and applicability towards many meta-learning algorithms.
Note that training a predictor constitutes a task related to the NAS task, but fundamentally a different one -- the former is about approximating a performance metric $s$, the latter about optimising it.
Also, because a set of NAS data $\mathcal{B}_{\mathcal{T}}$ directly defines a related predictor training task, we will use those terms interchangeably.

\noindent \textbf{Unified NAS Data.}
For meta-learning we assume we have $T$ NAS tasks with the corresponding NAS data, this might come from previous NAS trainings or NAS benchmarks.
We will refer to this set of available NAS data as $\Omega = \{ \mathcal{B}_{\mathcal{T}_i} \}_{i=0}^T$.


\subsection{Meta-learning NAS Performance Predictor}
We want to leverage the previous neural network training information to optimize the GCN to efficiently perform few-shot adaptation to new tasks, by utilising model-agnostic meta-learning.
Specifically, we consider predictor-training tasks (Eq.~\ref{eq:predictor_objective} with different $\mathcal{B}_{\mathcal{T}_i} \in \Omega$) to be independent tasks for the purpose of meta-learning -- we will refer to those tasks as \textit{meta-learning tasks}, to distinguish them from other tasks that we discuss on this paper.
To meta-learn our predictor, at each iteration $i$, we first randomly sample a subset of $K$ meta-learning tasks from, $\{ \mathcal{B}^i_j \}_{j=0}^K \subset \Omega$, where $\mathcal{B}^i_j$ is its $j$-th sampled task in iteration $i$. 

From each meta-learning tasks $\mathcal{B}^i_j$ we randomly sample $N_{finetune}$ and $N_{val}$ unique architecture-performance pairs to form the fine-tuning and validation examples for the task respectively. It is important to note as sampled from the same $\mathcal{B}^i_j$, they contain performance from the same NAS task (also dataset) and are enforced to be mutually exclusive:
\begin{gather}
    \mathcal{F}^{i}_{j} = \{ x_n \sim \mathcal{B}^i_j \}_{n=0}^{N_{finetune}} \\
    \mathcal{V}^{i}_{j} = \{ x_n \sim \mathcal{B}^i_j \setminus \mathcal{F}^{i}_{j} \}_{n=0}^{N_{val}}
\end{gather}
where $x \sim \mathcal{X}$ denotes an element $x$ sampled uniformly from the set $\mathcal{X}$.
In few-shot learning, $N_{finetune}$ and $N_{val}$ are generally a very small number of datapoints compared to the total number of datapoints in $\Omega$.

In order to meta-learn the predictor to quickly adapt given a small set of finetuning datapoints, lets assume at the beginning of iteration $i$ we start from an initialization $\theta^i$.
A step of the MAML inner loop on each finetuning task $\mathcal{F}^{i}_{j}$ is then taken to minimize the expected loss on this task using gradient descent, obtaining task-specific parameters $\theta^i_j$:
\begin{equation} \label{eq-maml-inner-loop}
  \theta^{i}_{j} = \theta^{i} - \alpha \nabla_{\theta} \mathbb{E}_{(A,s) \sim \mathcal{F}^i_{j}} \mathcal{L}(f_{\theta^{i}}(A); s),
\end{equation}
where $\alpha$ is the inner loop learning rate.
We further use the set of selected validation points for each task, $\mathcal{V}^{i}_{j}$, to test if the fine-tuning of the predictor worked and optimize $\theta$ in the outer loop to provide a good initialization for fine-tuning.
In MAML, this can be seen as gradually optimizing $\theta$ such that the expected loss is minimized after a few gradient steps across the entire set $\Omega$.
To update the initalization $\theta^i$, we perform a step of the outer loop with learning rate $\beta$ via:
\begin{equation} \label{eq-maml-outer-loop}
\textstyle  \theta^{i+1} \leftarrow \theta^{i} - \beta \nabla_\theta \sum_{j} \mathbb{E}_{(A,s) \sim \mathcal{V}^{i}_{j}} \mathcal{L}(f_{\theta^{i}_j}(A); s).
\end{equation}
In our case, we use a modified version of MAML called Body Only Meta Learning (BOIL) which updates all GCN parameters except the final layer in the inner loop and all parameters in the outer loop. This enforces the GCN to learn new graph features when trained on architecture-performance pairs on a new dataset instead of re-using the features \cite{oh2020boil}. We observe empirically in section \ref{sec:tb101}, BOIL performs better than MAML \cite{finn2017model} and ANIL \cite{raghu2019rapid} and provides better out-of-domain generalization. The inner-loop and outer-loop updates for BOIL can be written as follows:
\begin{align}
  \phi^{i}_j &= \phi^i - \alpha \nabla_\phi \mathbb{E}_{(A,s) \sim \mathcal{F}^i_j} \mathcal{L}(f_{\theta^i}(A); s) \\
  \textstyle  \theta^{i+1} &\leftarrow \theta^i - \beta \nabla_\theta \sum_{j} \mathbb{E}_{(A,s) \sim \mathcal{V}^{i}_{j}} \mathcal{L}(f_{[\phi^i_j, \lambda^i]}(A); s).
\end{align}
where the GCN parameters $\theta$ are divided into non-final layer and final layer parameters: $\theta = [\phi, \lambda]$.

Finally after meta-learning our predictor using set of previous neural network training and benchmarks ($\Omega$) we apply it to an unseen NAS task $\mathcal{T}_{diverse}$ and try to predict the performance of (possibly completely new) NN architectures when trained and evaluated on this new dataset. We do so in a predictor-based NAS setup, starting with zero-shot predictions to choose the model and fine-tuning our predictor every time we evaluate a NN architecture. The exact workflow of our method can be understood in Figure \ref{fig:overview}.

\noindent \textbf{Meta-testing with fine-tuning: }It is observed fine-tuning on top of meta-learning boosts the predictive performance further and helps generalize to unseen domains \cite{hu2022pushing}. To this end while fine-tuning on held-out dataset we lower the learning rate and perform a grid search to choose the optimal number of inner iterations from the set $\{5, 10, 20, 50, 100\}$. We choose the number of inner iterations that maximize the cross-validation Spearman correlation on the fine-tuning set. The boost in predcitor's performance is visualized in Figure \ref{fig:abl2}.

\subsection{Unifying Cell-based NAS Benchmarks}
\label{sec:unified}
In order to test our method for transferability from one search space to another we consider a total of 14 datasets over four cell-based search spaces, namely NAS-Bench-101, NAS-Bench-201, TransNAS-Bench-101-Micro and NAS-Bench-ASR. We outline the details of the datasets and the corresponding search spaces in table \ref{tab:unified_list} and appendix \ref{app:a}. We restrict ourselves to cell-based search spaces as they share some operations (such as \textit{Conv1} and \textit{Conv3} between NB101 and NB201) and thus can benefit from the transfer learning. Macro search spaces on the other hand are more dissimilar to each other - with almost no information common among these search spaces, they are bound to less advantage in transfer learning.

\begin{wraptable}{rh}{7.5cm}
\footnotesize
\setlength{\tabcolsep}{3.5pt}
\centering
\begin{tabular}{llcl}
\hline
NAS-Bench & Dataset & \# Datapoints & Metric  \\ \hline
NB101 & CIFAR10 & 423k & Acc.  \\
NB201 & CIFAR10 & 15.6k & Acc. \\
NB201 & CIFAR100 & 15.6k & Acc. \\
NB201 & ImageNet & 15.6k & Acc. \\
NB360 & NinaPro & 15.6k & Acc. \\
NB360 & Darcy Flow & 15.6k & Rel. $\mathcal{L}_2$ \\
LatBench & CPU latency & 15.6k & Latency \\
LatBench & GPU latency & 15.6k & Latency \\
TB101 & Object Clas. & 4k & Acc.  \\
TB101 & Scene Clas. & 4k & Acc. \\
TB101 & Room Layout & 4k & MSE Loss \\
TB101 & Jigsaw & 4k & Acc. \\
TB101 & Autoencoding & 4k & SSIM \\
TB101 & Surface Normal & 4k & SSIM \\
TB101 & Sem. Segment. & 4k & Acc. \\
NB-ASR & TIMIT & 8.2k & PER\\ \hline
\end{tabular}
\caption{List of datasets and NAS-Benchmarks unified to meta-learn the predictor.}
\label{tab:unified_list}
\end{wraptable}

We represent a DAG as set of node features and adjacency matrix, $\mathcal{G} = (V, A)$ where $V$ is node feature matrix containing one-hot representation of operations happening at a given node and $A$ is the adjacency matrix of the graph. In order for the GCN to operate on graphs of different sizes we need to unify their representations. We do so by representing the node feature matrices under the union of operations across all search spaces.
For example, let's consider two search spaces $S_1$ and $S_2$ having three operations (e.g., \textit{Conv3}, \textit{Conv1} and \textit{AvgPool}) and two operations (e.g., \textit{Conv3} and \textit{MaxPool}), respectively. A graph $G_1$ from $S_1$ will be represented via its respective adjacency matrix and one-hot encodings of the four unified node operations.
%
Therefore, while unifying NB101, NB201, TB101 and NB-ASR we end up with $11$ operations, including a skip-connection, a zeroize, a linear, 6 convolution, and 2 pooling  operations. Along with these we also include a node each for input and output as well as a global node. Thus, our unified search space has overall $14$ node operations.

As the operations in the unified representation of search spaces are more than the operations in its individual search spaces, we construct a unified search space which leverages and allows for any operation among the union of operations across individual search spaces. We name this strategy \textit{mixed-ops}, as we allow mixing operations from different search space. \textit{Mixed-ops} increases the search space of possible architectures to 2.35 billion, $\approx 5000 \times$ higher than the sum of architectures in individual search spaces. We use this to search for optimal neural architectures on NB360 tasks and more details can be found in Appendix \ref{app:a}.


\section{Experiments}
We conduct extensive experiments to validate our framework. First, we focus on a set of synthetic datasets to understand the transferability aspects of our method, particulary how quality and quantity of meta-training set affects our NAS results (section \ref{sec:synthetic}). Then we look at the TB101 search space to evaluate our predictor for cross-dataset transferability (same search space but different dataset). We also investigate various components of our algorithm through careful ablation analysis (section \ref{sec:tb101}). Secondly, we focus on the NB201 search space to evaluate our predictor's performance on both cross-dataset and cross-search space (different search space and dataset) tasks (section \ref{sec:nb201}). Finally, we highlight the use of unified search space and performing meta-learning on this search space to give superior results on five NB360 tasks (section \ref{sec:nb360}).

In order to evaluate the transfer ability of our predictor in these search spaces, we compare the Spearman correlation achieved by the performance predictor on held-out testing dataset. We also compare previous NAS methods directly by looking at the performance of the best found model as a function of search time. Additionally, a full set of experimental details required to reproduce all our results can be found in Appendix \ref{app:b}.
\subsection{Synthetic Experiments}
\label{sec:synthetic}

\begin{figure*}[t]
    \centering
    \includegraphics[width=.95\textwidth]{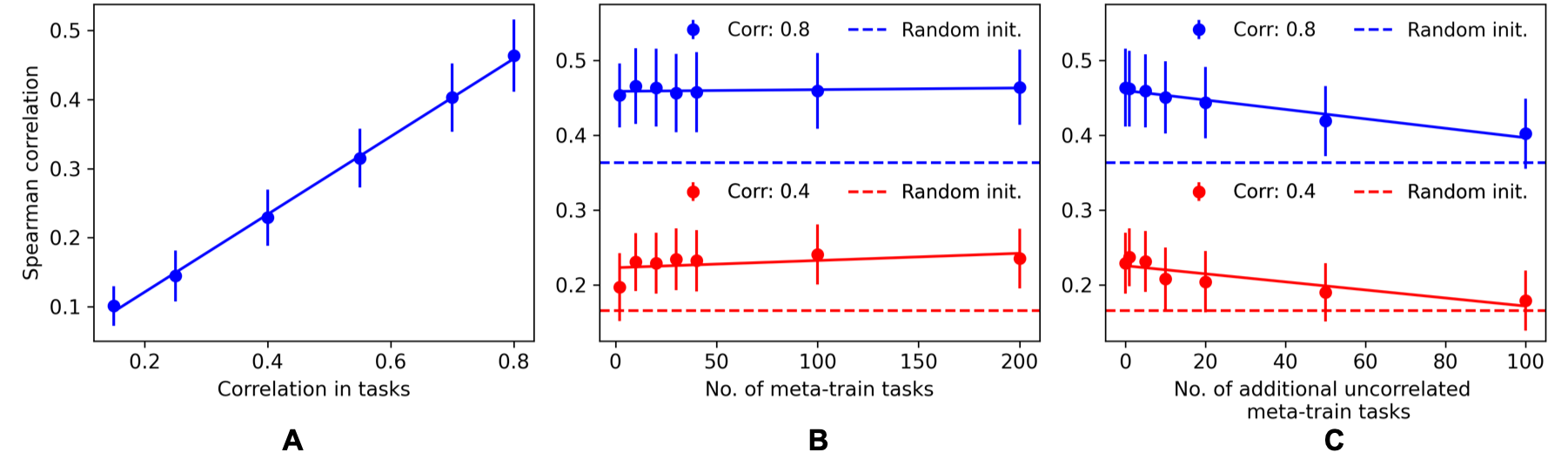}
    \caption{Performance of \mlnas on synthetic experiments to understand the dependence of the predictor on \textbf{(A)} Increasing the correlation between tasks, \textbf{(B)} Number of meta-training tasks, \textbf{(C)} Number of un-correlated tasks in meta-training. The error bars represent standard errors calculated by 40 independent runs.}
    \label{fig:syn}
\end{figure*}

We devise a systematic study to understand the transferability aspects of our predictor by simulating a number of synthetic accuracy values for architectures from NB201. We simulate multiple meta-learning tasks by corrupting the validation accuracy of architectures from NB201 on CIFAR-100. We do so by adding a fixed white Gaussian noise:
\begin{equation}
    s_{task_j} = s_{C100} + \epsilon_{j} ; \hspace{3mm} \epsilon_{j} \sim \mathcal{N}(\vec{0}, \sigma^2I)
\end{equation}
Where, $s_{C100}$ is the vector of CIFAR-100 accuracies on NB201 architectures, $s_{task_j}$ is the simulated accuracy for $j^{th}$ task and $\sigma^2$ controls the correlation we observe between meta-learning tasks. We perform a leave-one-out validation where we meta-train on all but the last task and meta-test on the last task. In order to understand different aspects on which the robustness of the predictor depends, a) we vary the correlation between the meta-training and meta-testing tasks, b) number of meta-training tasks and c) the number of uncorrelated tasks in meta-training. We meta-train our predictor using 256 architectures, finetune using 5 architectures and test on the remaining architectures. We visualize our experiments in Figure \ref{fig:syn} and summarize our main findings as follows:

\textbf{Correlation between meta-training tasks:} Increasing the correlation between the meta-training tasks and meta-test tasks (for a fixed number of meta-training tasks) leads to better performance, as the meta-test set is less out-of-distribution. This implies, having better quality tasks will almost surely improve the predictor.

\textbf{Number of meta-training tasks:} Increasing the number of meta-training tasks while keeping the correlation between the tasks fixed doesn't affect the Spearman correlation significantly. We observe a slight increase in Spearman correlation from 1 meta-training task to 5 meta-training tasks (likely explained by meta-overfitting) but there is no significant change ($<\pm 2\%$) if we increase the number of meta-training tasks further. This leads us to conclude, even having a modest number of meta-training tasks like few previous NAS benchmarks we use, is sufficient for improved predictor performance.

\textbf{Number of uncorrelated tasks:} Does including a few noise tasks on top of 10 correlated tasks in meta-training set affect the performance? We observe adding these uncorrelated tasks in the meta-learning deteriorates the transfer ability of our method. Having 10x more uncorrelated tasks in meta-training makes the meta-learner under-fit, eventually converging to the performance of a randomly initialized predictor. Therefore, one of the ways to improve transferability of NAS predictor is ensuring lack of extremely uncorrelated tasks while meta-training. We observe meta-training using previous NAS benchmarks provides with an excellent set of correlated tasks for meta-learning.

\subsection{TransNAS-Bench-101 Search Space}
\label{sec:tb101}

\begin{table}
\footnotesize
\centering
\setlength{\tabcolsep}{3pt}
\begin{minipage}[t]{0.45\columnwidth}%
    \scalebox{0.8}{
    \begin{tabular}{@{}lccc@{}}
        \toprule
        \textbf{Dataset} & \textbf{Rand. Init.}  & \textbf{Naive Tran.} & \textbf{\mlnas} \\
        (\# models) & 5 & 5 & 5\\ \midrule
        Scene Clas.    & 0.66$_{\pm 0.06}$ & \secondbest{0.80}$_{\pm 0.02}$ & \best{0.88}$_{\pm 0.01}$ \\
        Object Clas.   & 0.47$_{\pm 0.07}$ & \secondbest{0.73}$_{\pm 0.01}$ & \best{0.78}$_{\pm 0.02}$ \\
        Jigsaw         & 0.51$_{\pm 0.05}$ & \secondbest{0.71}$_{\pm 0.02}$ & \best{0.77}$_{\pm 0.02}$ \\
        Room Layout    & 0.35$_{\pm 0.08}$ & \secondbest{0.73}$_{\pm 0.02}$ & \best{0.79}$_{\pm 0.05}$ \\
        Sem. Segment.  & 0.70$_{\pm 0.05}$ & \secondbest{0.77}$_{\pm 0.03}$ & \best{0.83}$_{\pm 0.03}$ \\
        Surface Normal & 0.60$_{\pm 0.09}$ & \secondbest{0.80}$_{\pm 0.02}$ & \best{0.81}$_{\pm 0.01}$ \\
        Autoencoding   & 0.11$_{\pm 0.04}$ & \best{0.43}$_{\pm 0.03}$ & \secondbest{0.42}$_{\pm 0.05}$ \\ \midrule
        Overall Mean   & 0.49$_{\pm 0.05}$ & \secondbest{0.71}$_{\pm 0.03}$ & \best{0.75}$_{\pm 0.03}$ \\ \bottomrule
    \end{tabular}}
    \caption{Spearman-$\rho$ on cross-dataset transfer within TB101, performed using leave-one-out validation. Each tabular entry represents the mean Spearman-$\rho$ and standard errors.}
    \label{tab:tb101}
\end{minipage}%
\hfill
\begin{minipage}[t]{.53\columnwidth}%
    \scalebox{0.8}{
    \begin{tabular}{@{}lccccc@{}}
        \toprule
        \textbf{Dataset} & \multicolumn{4}{c}{ \textbf{\mlnas} } & \textbf{Rand. Init.} \\
        (\# models) & 0 & 5 & 25 & 50 & 50 \\ \midrule
        Scene Clas.    & 0.76$_{\pm 0.00}$ & \secondbest{0.88}$_{\pm 0.01}$ & \secondbest{0.89}$_{\pm 0.04}$ & \best{0.91}$_{\pm 0.00}$ & 0.81$_{\pm 0.01}$ \\
        Object Clas.   & \secondbest{0.78}$_{\pm 0.00}$ & \secondbest{0.78}$_{\pm 0.02}$ & \secondbest{0.82}$_{\pm 0.01}$ & \best{0.83}$_{\pm 0.00}$ & 0.66$_{\pm 0.01}$ \\
        Jigsaw         & \secondbest{0.76}$_{\pm 0.00}$ & \secondbest{0.77}$_{\pm 0.02}$ & \secondbest{0.79}$_{\pm 0.00}$ & \best{0.8}$_{\pm 0.00}$ & 0.63$_{\pm 0.01}$ \\
        Room Layout    & 0.64$_{\pm 0.00}$ & \secondbest{0.79}$_{\pm 0.05}$ & \secondbest{0.92}$_{\pm 0.01}$ & \best{0.93}$_{\pm 0.00}$ & 0.66$_{\pm 0.03}$ \\
        Sem. Segment.  & 0.62$_{\pm 0.00}$ & \secondbest{0.83}$_{\pm 0.03}$ & \secondbest{0.91}$_{\pm 0.01}$ & \best{0.92}$_{\pm 0.00}$ & 0.81$_{\pm 0.00}$ \\
        Surface Normal & 0.74$_{\pm 0.00}$ & 0.81$_{\pm 0.01}$ & 0.81$_{\pm 0.00}$ & \best{0.84}$_{\pm 0.01}$ & 0.81$_{\pm 0.01}$ \\
        Autoencoding   & 0.28$_{\pm 0.00}$ & \secondbest{0.42}$_{\pm 0.05}$ & \secondbest{0.68}$_{\pm 0.00}$ & \best{0.76}$_{\pm 0.00}$ & 0.32$_{\pm 0.04}$ \\ \midrule
        Overall Mean   & 0.66$_{\pm 0.00}$ & \secondbest{0.75}$_{\pm 0.03}$ & \secondbest{0.83}$_{\pm 0.00}$ & \best{0.86}$_{\pm 0.00}$ & 0.67$_{\pm 0.00}$ \\ \bottomrule
    \end{tabular}}
    \caption{Increasing the number of fine-tuning models improves the Spearman-$\rho$ on cross-dataset transfer within TB101. Entries in blue highlight improvements over Rand. Init. despite using less data. }
    \label{tab:abl1}
\end{minipage}
\end{table}

TransNAS-Bench-101 (TB101) is a NAS benchmark developed for multi-task NAS methods, especially to experiment with different transfer learning methods for NAS. TB101 has seven computer vision tasks spanning classification, regression, pixel-wise prediction, and self-supervised tasks. It has 4096 micro (or cell-based) and 3256 macro search space architectures. We focus on cell-based search space to leverage the similarity between them.

\begin{wrapfigure}{r}{0.5\textwidth}
\centering
\includegraphics[width=0.5\textwidth]{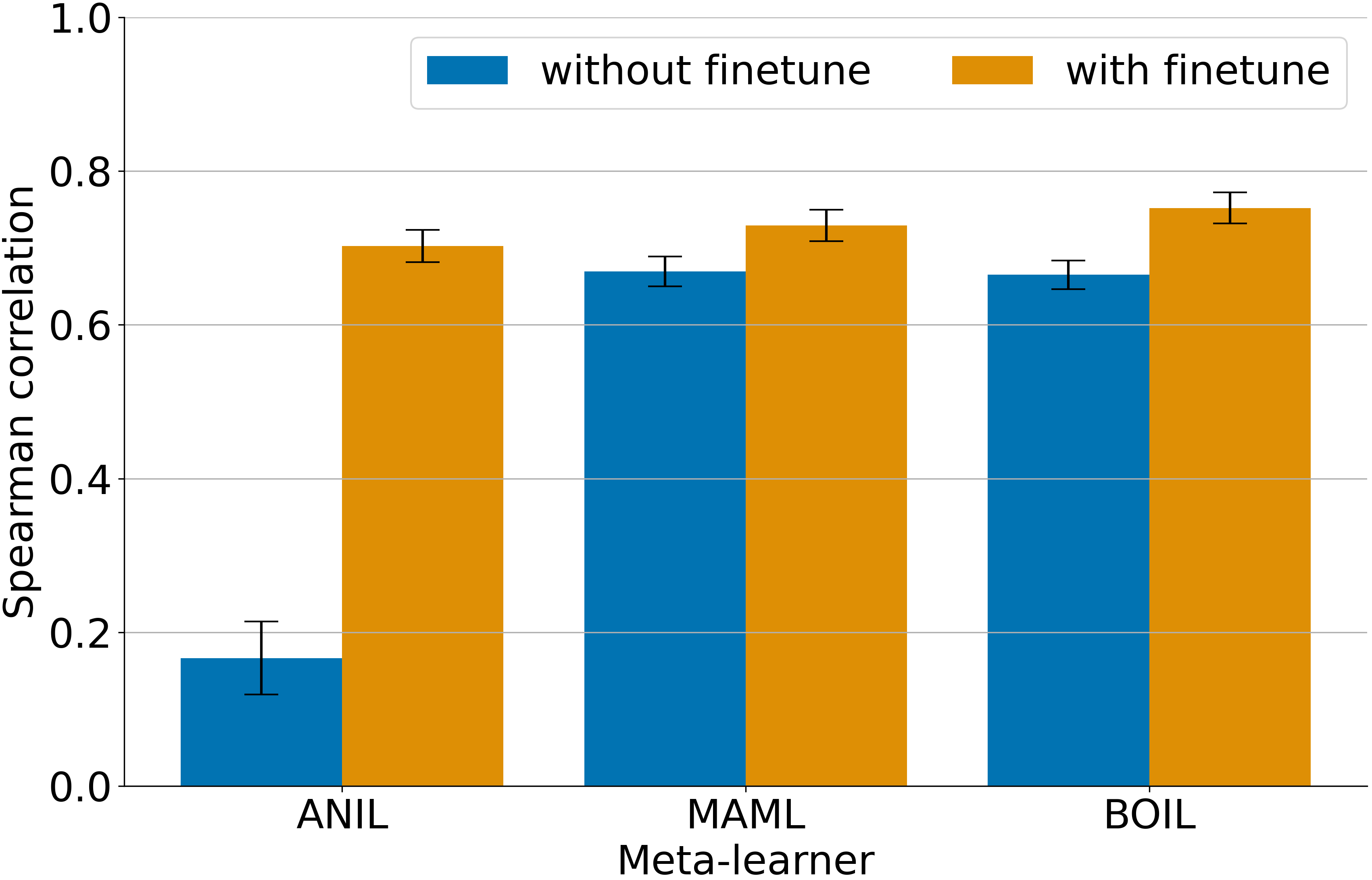}
\caption{Performance of different MAML algorithms on cross-dataset transferability within TB101. Error-bars correspond to standard errors.}
\label{fig:abl2}
\end{wrapfigure}

We perform a leave-one-out validation scheme to evaluate our transfer learning ability on each one of the seven tasks in TB101. We meta-learn using 2048 architectures with six dataset accuracy values, fine-tune using 5 architectures with held-out dataset accuracy values, and test for performance prediction on the held-out dataset for the remaining 2043 architectures. Firstly, we compare the initialization achieved using meta-learning against a randomly initialized performance predictor in Table \ref{tab:tb101}. We observe meta-initialization followed by fine-tuning using 5 datapoints gives similar Spearman correlation to random initialization followed by training using 50 datapoints. Being able to achieve high Spearman correlation given a small set of datapoints is important in NAS, as generating each datapoint (architecture-performance pair) requires hours of GPU training time and is a bottleneck in many NAS methods.

We also compare our meta-learnt NAS against a more traditional transfer learning approach of pre-training the predictor on one of the datasets and fine-tuning on the target dataset. Specifically, we train on the 2048 architectures from six training datasets seperately, followed by fine-tuning using 5 datapoints from the target dataset and then averaging the prediction Spearman correlation achieved using each training dataset. This is a very strong baseline, especially with NAS transferability but we are able to surpass it signifying the metalearnt initialization is able to learn more complicated transfer learners than simply pre-training+fine-tuning the predictor.



\subsubsection{Ablation analysis}

\textbf{Increasing the number of fine-tuning models: }An important condition to be met by the performance predictor is when supplied with more datapoints i.e. architecture-accuracy pairs it should almost surely output more accurate estimates of the performance. We observe this trend when we assign more fine-tuning models in the held-out evaluation, signifying the performance predictor benefit from fine-tuning and improving its predictions (Table \ref{tab:abl1}).

\textbf{Different model-agnostic meta-learners: }We experiment with different model-agnostic meta-learning algorithms for their transfer learning abilities in TB101. We compare MAML \cite{finn2017model}, BOIL \cite{oh2020boil} and ANIL \cite{raghu2019rapid} in Figure \ref{fig:abl2}. We also observe while meta-testing, fine-tuning with a lower learning rate is beneficial, consistent with previous work \cite{hu2022pushing}.

\subsection{NAS-Bench-201 Search Space}

\label{sec:nb201}
NAS-Bench-201 (NB201) is one of the most popular NAS benchmarks for NAS evaluation, it provides training information for three classification datasets namely CIFAR-10, CIFAR-100 \cite{krizhevsky2009learning} and ImageNet-16-120 \cite{chrabaszcz2017downsampled} on 15,625 architectures from a cell-based search space. We append four more tasks to the original NB201, with two from NB360 \cite{tu2021bench} and two from LatBench \cite{dudziak2020brpnas}. The two NB360 tasks are: PDE solver using Darcy Flow dataset \cite{li2020fourier} and prosethetic control using Ninapro DB5 \cite{atzori2012building}. Whereas the two LatBench tasks record latency values on desktop CPU (core-i7-7820x-fp32) and embedded GPU (jetson-nono-fp16) for the NB201 search space architectures when trained on CIFAR-100. We visualize the correlation in the accuracy/latency values across the datasets in Appendix Figure \ref{fig:corr}.

\begin{table}
\footnotesize
\centering
\setlength{\tabcolsep}{3pt}
\begin{minipage}[t]{0.48\columnwidth}%
    \scalebox{0.9}{
    \begin{tabular}{@{}lccc@{}}
    \toprule
    \textbf{Dataset} & { \textbf{Rand. Init.} }  & \textbf{Naive Tran.} & \textbf{\mlnas} \\
    (\# models) & 5 & 5 & 5\\ \midrule
    CIFAR10         & 0.6$_{\pm 0.08}$  & 0.73$_{\pm 0.03}$ & \best{0.92}$_{\pm 0.01}$ \\
    CIFAR100        & 0.61$_{\pm 0.07}$ & 0.73$_{\pm 0.03}$ & \best{0.93}$_{\pm 0.01}$ \\
    ImageNet        & 0.6$_{\pm 0.07}$  & 0.74$_{\pm 0.03}$ & \best{0.89}$_{\pm 0.01}$ \\
    Ninapro         & 0.13$_{\pm 0.04}$ & 0.2$_{\pm 0.04}$  & \best{0.32}$_{\pm 0.03}$ \\
    Darcy Flow      & 0.67$_{\pm 0.06}$ & 0.64$_{\pm 0.03}$ & \best{0.73}$_{\pm 0.03}$ \\
    CPU Latency     & 0.73$_{\pm 0.09}$ & 0.62$_{\pm 0.05}$ & \best{0.8}$_{\pm 0.03}$ \\
    GPU Latency     & \best{0.77}$_{\pm 0.06}$ & 0.67$_{\pm 0.03}$ & 0.73$_{\pm 0.04}$ \\ \midrule
    Overall Mean    & 0.59$_{\pm 0.04}$ & 0.62$_{\pm 0.04}$ & \best{0.75}$_{\pm 0.03}$ \\ \bottomrule
    \end{tabular}}
    \caption{Spearman-$\rho$ on cross-dataset transfer within NB201, performed using leave-one-out validation. Each tabular entry represents the mean Spearman-$\rho$ and standard errors.}
    \label{tab:nb201}
\end{minipage}%
\hfill
\begin{minipage}[t]{0.50\columnwidth}%
    \raisebox{4mm}{\scalebox{0.9}{
    \begin{tabular}{@{}lcccc@{}}
    \toprule
    \multirow{2}{*}{\textbf{Dataset}} & \multicolumn{2}{c}{ \textbf{Predictor Perf.} }  & \multicolumn{2}{c}{\textbf{NAS Perf. (\%)}} \\ \cmidrule(r){2-3} \cmidrule(l){4-5}
     & Meta-D2A & \mlnas & Meta-D2A & \mlnas\\ \midrule
    CIFAR10 & 0.56 & \best{0.77} & \best{94.37}$_{\pm 0.00}$ & \best{94.37}$_{\pm 0.02}$ \\
    CIFAR100 & 0.58 & \best{0.76} & \best{73.51}$_{\pm 0.00}$ & 73.49$_{\pm 0.08}$ \\
    ImageNet & 0.54 & \best{0.78} & 46.9$_{\pm 0.10}$  & \best{47.2}$_{\pm 0.06}$ \\
    Ninapro & \best{0.27} & 0.2  & 92.84$_{\pm 0.06}$  & \best{92.85}$_{\pm 0.2}$ \\ \midrule
    Overall Mean & 0.49 & \best{0.63} & 76.9 & \best{77.0} \\ \bottomrule
    \end{tabular}}}
    \caption{Predictor and NAS performance of \mlnas vs. Meta-D2A on NB201 search space. Predictor performance (Spearman-$\rho$) obtained through zero-shot (i.e. no fine-tuning) and NAS performance (the best accuracy) achieved after 30 queries (averaged over 10 runs).}
    \label{tab:metad2a}
\end{minipage}
\end{table}

\noindent \textbf{Cross dataset transferability: } We perform a leave-one-out validation scheme similar to section \ref{sec:tb101} to assess the transferability within the same search space. We meta-learn our performance predictor using 7,812 architectures and six datasets, fine-tune using 5 architectures with the held-out dataset accuracy/latency values and then, finally test on the remaining 7,808 architectures. In Table \ref{tab:nb201}, we compare \mlnas against randomly initializing the predictor and the traditional transfer learning approach of pretraining on a dataset and fine-tuning on the target dataset. We observe \mlnas significantly outperforms the naive transfer learning benchmark and is on par with randomly initialized predictors with $10 \times$ more training datapoints.


\noindent \textbf{Comparison with MetaD2A: }In Table \ref{tab:metad2a} we compare our predictor (meta-learnt using cross-dataset transfer) and overall NAS performance to MetaD2A \cite{lee2021rapid}, a previously published NAS algorithm that also uses meta-learning to prevent NAS re-training. We only consider CIFAR10, CIFAR100, ImageNet-16-120 and Ninapro datasets as MetaD2A is restricted to classification tasks. We use the default parameters for the MetaD2A generator and predictor and choose the best model based on validation accuracy after 30 queries for both MetaD2A and MP-NAS.


\noindent \textbf{Cross search space transferability: }We also evaluate our method's transfer ability for out-of-domain generalization to a completely new search space. As far as we know this is the first study to establish and benchmark cross-search space transferability -- owing to large differences in graph structures, operations, macro-architectures and hyperparameters between search spaces. We use a unified representation of four cell-based search spaces as described in section \ref{sec:unified} to meta-learn and test our predictor. We observe even without accounting for differences in macro-architectures and hyperparameters used during training, the knowledge about which graph features are important is transferable across search spaces.

We meta-learn using 7,812 architectures and seven previously considered datasets from NB201 search space followed by fine-tuning using 5 architectures on the new dataset and search space and finally, testing on the rest of the architectures in the new search space. The new search spaces we consider are TB101-micro (which has same graph structure as NB201 with one operation missing), NB-101 (which has different graph structure and 2 more operations than NB201) and NB-ASR (which not only has a different graph structure and uses different operations, but is also proposed for a completely different modality). We observe the transfer ability reduces as we go from TB101 to NB101 and NB-ASR, as they are more out-of-distribution. Nonetheless, we are still better than random initialization (Table \ref{tab:nb201_cross}).


\begin{table}
\footnotesize
\centering
\setlength{\tabcolsep}{3pt}
\begin{minipage}[t]{0.45\textwidth}
    \scalebox{0.8}{
    \begin{tabular}{@{}lccc@{}}
        \toprule
        \textbf{SS/Dataset} & \textbf{Rand. Init.} & \textbf{Naive Tran.} & \textbf{\mlnas} \\
        (\# models)    & 5 & 5 & 5\\ \midrule
        TB/Scene Clas.      & 0.66$_{\pm 0.06}$ & 0.76$_{\pm 0.03}$ & \best{0.87}$_{\pm 0.01}$ \\
        TB/Object Clas.     & 0.47$_{\pm 0.07}$ & 0.63$_{\pm 0.03}$ & \best{0.74}$_{\pm 0.02}$ \\
        TB/Jigsaw           & 0.51$_{\pm 0.05}$ & 0.62$_{\pm 0.03}$ & \best{0.74}$_{\pm 0.02}$ \\
        TB/Room Layout      & 0.35$_{\pm 0.08}$ & 0.59$_{\pm 0.04}$ & \best{0.72}$_{\pm 0.05}$ \\
        TB/Sem. Segment.    & 0.70$_{\pm 0.05}$ & 0.74$_{\pm 0.03}$ & \best{0.83}$_{\pm 0.03}$ \\
        TB/Surface Normal   & 0.60$_{\pm 0.09}$ & 0.72$_{\pm 0.03}$ & \secondbest{0.79}$_{\pm 0.01}$ \\
        TB/Autoencoding     & 0.11$_{\pm 0.04}$ & \secondbest{0.4}$_{\pm 0.04}$ & \best{0.47}$_{\pm 0.05}$ \\
        NB101/CIFAR10       & -0.06$_{\pm 0.06}$ & \secondbest{0.4}$_{\pm 0.04}$ & \best{0.46}$_{\pm 0.04}$ \\
        NB-ASR/Timit        & -0.11$_{\pm 0.05}$ & \secondbest{0.02}$_{\pm 0.02}$ & \best{0.13}$_{\pm 0.04}$ \\ \midrule
        Overall Mean        & 0.35$_{\pm 0.05}$ & \secondbest{0.54}$_{\pm 0.02}$ & \best{0.64}$_{\pm 0.03}$ \\ \bottomrule
    \end{tabular}}
    \caption{Spearman-$\rho$ on cross search space transfer from NB201 to TB101, NB101 and NB-ASR. Each tabular entry represents the mean Spearman-$\rho$ and standard errors.}
    \label{tab:nb201_cross}
\end{minipage}
\hfill
\begin{minipage}[t]{0.53\textwidth}
    \scalebox{0.8}{
    \begin{tabular}{@{}lccccc@{}}
        \toprule
        \textbf{Dataset (Metric)} & \textbf{WRN} & \textbf{DASH} & \textbf{Expert} & \textbf{Rand. Pred.} & \textbf{\mlnas} \\ \midrule
        Spherical (Acc.) & 14.23 & 28.72 & \secondbest{32.59} & 31.47 & \best{35.6}\\ 
        Darcy Flow ($L_2$) & 0.073 & \secondbest{0.0079} & 0.008 & 0.010 & \best{0.0078} \\ 
        NinaPro (Acc.) & \secondbest{93.22} & \best{93.4} & 91.27 & 89.78 & 92.71 \\ 
        ECG (F1) & 0.57 & 0.68 & \best{0.72} & 0.68 & \secondbest{0.68} \\ 
        Satellite (Acc.) & 84.51 & \best{87.72} & 80.2 & 86.04 & \secondbest{87.33} \\ 
        \bottomrule
    \end{tabular}}
    \caption{Performance of \mlnas across five diverse tasks selected from NAS-Bench-360. \mlnas outperforms hand-designed expert models on 4/5 tasks and DASH on 3/5 tasks.}
    \label{tab:nb360}
\end{minipage}
\end{table}

\subsection{NAS-Bench-360 Evaluation}
\label{sec:nb360}


NAS-Bench-360 is a recently proposed benchmark suite to evaluate NAS algorithms on diverse domains. We evaluate and compare \mlnas against previous baselines on five NB360 NAS tasks. In particular we consider Spherical CIFAR-100 \cite{cohen2018spherical}, Darcy Flow \cite{li2020fourier}, Ninapro DB5 \cite{atzori2012building}, ECG \cite{clifford2017af} and Satellite \cite{petitjean2012satellite} datasets and compare \textbf{\mlnas} against wide-resnet (\textbf{WRN}), expert crafted architecture (\textbf{Expert}) and a recently published NAS method for diverse task, \textbf{DASH} \cite{shen2022efficient}.
We meta-trained the predictor on all four benchmarks considered in this work, and also perform the search with randomly initialised predictor (\textbf{Rand. Pred.}) to evaluate the efficacy of meta-learning directly. We run the predictor-based NAS for 20 steps, re-training the predictor every 4 steps and choosing the top model based on predictions for 10,000 randomly sampled architectures every step. It is worth noting that even with a simple experimental setup like ours, MP-NAS outperforms hand-designed expert models on four tasks and DASH on three tasks out of total five tasks. More details about datasets, hyperparameters and comparison with randomly initialized predictor can be found in Appendix \ref{app:c}.







In this paper, we propose MP-NAS, a novel NAS algorithm that leverages past neural network training by meta-initializing the performance predictor for faster NAS search, especially in diverse tasks. By demonstrating cross-search space transferability and leveraging meta-learning on a unified search space we perform NAS and beat previous NAS baselines on 3/5 NAS-Bench-360 tasks. One limitation of the proposed \mlnas is that it leverages similarity of graph structure or operations for transferability between search spaces. However, many search spaces are disjoint to each other and have been designed with a particular task in mind. Like most macro search spaces including TB101-macro \cite{duan2021transnas} and BLOX \cite{chau2022blox} or extremely large search spaces like DARTS \cite{liu2018darts} and NB301 \cite{siems2020bench} designing a unified search space is rather difficult and isn't supported in its current form. For future work, it would be interesting to improve interpretability of the learned graph feature representations, and extend our method to unify more NAS benchmarks and search spaces, including macro-architecture and surrogate search spaces.

\section{Conclusion, Limitations and Future Work}
\textbf{Conclusion: }We propose MP-NAS, a novel NAS algorithm that leverages past neural network training by meta-initializing the performance predictor for faster NAS search, especially in diverse tasks. By demonstrating cross-search space transferability and leveraging meta-learning on a unified search space we perform NAS and beat previous NAS baselines on 3/5 NAS-Bench-360 tasks.

\textbf{Limitations: }\mlnas leverages similarity of graph structure or operations for transferability between search spaces. But many search spaces are disjoint to each other and have been designed with a particular task in mind. Like most macro search spaces including TB101-macro \cite{duan2021transnas} and BLOX \cite{chau2022blox} or extremely large search spaces like DARTS \cite{liu2018darts} and NB301 \cite{siems2020bench} designing a unified search space is rather difficult and isn't supported in its current form.

\textbf{Future work: }In its current form, \mlnas uses validation accuracy or latency from previous NAS Benchmarks on different datasets and search spaces to meta-learn the predictor. One of the future directions for \mlnas is designing meta-learning tasks such that it improves the performance of NAS on a downstream task. Recent success of zero-cost proxies \cite{abdelfattah2021zero-cost,mellor2021nwot,chen2021neural,ming2021zennas} or unsupervised task based pre-training \cite{liu2020labels, li2021generic} make ideal candidates to be included as the inner-loop tasks in our meta-learning pipeline.
Another avenue of future work is improving interpretability of the learned graph feature representations. Having the ability to explain the features after fine-tuning and meta-learning would help us in extending these findings and hand-crafting better neural architectures on a bigger search space. Having explainable features also means better transferability across domains.
Finally, we would like to extend our method to unify more NAS benchmarks and search spaces, including macro-architecture and surrogate search spaces.


\newpage

\balance
\bibliographystyle{unsrt}
\bibliography{main.bib}

\newpage
\appendix
\onecolumn

\section{Unified search space}
\label{app:a}

Our unified search space is constructed by intermixing four search spaces of existing NAS Benchmarks, including NB101, NB201, TB101 and NB-ASR.
These benchmarks comprise of overall 11 Operations: \textit{Conv1-D1}, \textit{Conv3-D1}, \textit{Conv5-D1}, \textit{Conv5-D2}, \textit{Conv7-D1}, \textit{Conv7-D2}, \textit{Linear}, \textit{AvgPool}, \textit{MaxPool}, \textit{Skip-Connection}, and \textit{Zeroize}. Most operations are incorporated in multiple benchmarks and these are detailed in Table~\ref{table:ops_unified}.

\begin{table}[h!]
\centering
\begin{tabular}{l|c|c|c|c}
\hline
\textbf{Operation} & \textbf{NB101} & \textbf{NB201} & \textbf{TB101} & \textbf{NB-ASR} \\
\hline
\textit{Conv1-D1} & \cmark & \cmark & \cmark & \xmark \\
\textit{Conv3-D1} & \cmark & \cmark & \cmark & \xmark \\
\textit{Conv5-D1} & \xmark & \xmark & \xmark & \cmark \\
\textit{Conv5-D2} & \xmark & \xmark & \xmark & \cmark \\
\textit{Conv7-D1} & \xmark & \xmark & \xmark & \cmark \\
\textit{Conv7-D2} & \xmark & \xmark & \xmark & \cmark \\
\textit{Linear} & \xmark & \xmark & \xmark & \cmark \\
\textit{AvgPool} & \xmark & \cmark & \xmark & \xmark \\
\textit{MaxPool} & \cmark & \xmark & \xmark & \xmark \\
\textit{Skip-Connection} & \xmark & \cmark & \cmark & \cmark \\
\textit{Zeroize} & \xmark & \cmark & \cmark & \cmark \\
\hline
\end{tabular}
\caption{Operations from unified search space that are covered in the four NAS Benchmarks (NB101, NB201, TB101 and NB-ASR). For a convolution operation the kernel size and dilation are indicated by the number following \textit{Conv}. For example, \textit{Conv5-D2} refers to convolution operation with kernel size of 5 and dilation equal to 2.}
\label{table:ops_unified}
\end{table}

\textbf{Mixed-Ops Extension: }
As discussed earlier in the paper, we extend each of the search space to leverage all unified operations so that we can quantify the potential of our proposed predictor.
Specifically, each search could generate graphs with node operations selected from a list of $11$ operations listed in Table \ref{table:ops_unified}. For example, NAS-Bench-201 which originally has $5$ operations and $6$ nodes would generate $15,625$ (i.e., $5^6$) cell graphs. However, after our extension this search space could use $11$ operations and thus generate $1,771,561$ (i.e., $11^6$) cell graphs.
Overall, the combination of all search spaces with our extension generate more than $2.35B$ cell graphs.

\textbf{Macro Architecture: }
In order to train models from the \textit{Mixed-Ops Extension} we fix the macro architecture for all the models. We use a macro architecture similar to previous search spaces (like NB101 and NB201) for all the architectures generated from this search spaces. This macro architecture comprises of a stem block, 3 stacks of cells connected by a residual block, and a classifier. The stem block consist of a $Conv3$ and $BatchNorm$. Each stack contains five cells with either 16, 32 or 64 output channels. Finally, the classifier comprises of a pooling and a linear layers and is adapted based on the number of classes in a dataset. Furthermore, depending on the dataset we transform all operations to either $1D$ or $2D$ operations to accordingly adapt the model.

\newpage

\section{Hyperparameters for predictor training}
\label{app:b}

Detailed information about hyperparameters used for meta-training our predictor are summarized in Table~\ref{tab:hyperparams}.

\begin{table}[h!]
    \centering
    \begin{tabular}{|c|c|}
    \hline
    \textbf{Hyperparameter} & \textbf{Setting description} \\ \hline
    Architecture & GCN \\
    Number of hidden layers & 4 \\
    Layer width & 600 \\
    Activation & ReLU \\
    Outer optimization algorithm & AdamW \cite{loshchilov2017decoupled} \\
    Outer learning rate & $8\times10^{-5}$\\
    Inner optimization algorithm & SGD \\
    Inner learning rate & 0.035 \\
    Number of inner steps & 6 \\
    Batch size & 64 \\
    Meta-learning epochs & 400 \\
    Dropout & 0.2 \\
    Number of meta-training runs & 1 \\
    Number of evaluation runs & 10 \\
    Total meta-training time & 18h 31m \\
    Computing infrastructure & Nvidia GTX 2080i \\
    \hline
    \end{tabular}
    \caption{List of hyperparameters used for meta-learning our predictor}
    \label{tab:hyperparams}
\end{table}

\section{NAS-Bench-360 evaluation}
\label{app:c}

To perform the NAS-Bench-360 evaluations on different datasets, we used the default hyperparameters for wide resnet from the DASH code-base (\url{https://github.com/sjunhongshen/DASH}). In Table \ref{tab:nb360_hyperparams}, we present the details of hyperparameters we used for training models per dataset. Most of these evaluations were performed on NVIDIA V100 GPUs.

We are able to consider five out of the ten datasets from NAS-Bench-360 owing to divergence issues with default hyperparameters in PSICOV, metric inconsistency between the code-base and the paper in Cosmic and DeepSEA and exceptionally slow dataloaders for CIFAR-10 and FSD-50k datasets. We believe the five datasets we consider still represent the diversity in the benchmark-suite and ensure a fair comparison.

\begin{table}[h!]
    \centering
    \begin{tabular}{|c|c|c|c|c|c|c|}
    \hline
    \textbf{Dataset} & \textbf{Batch size} & \textbf{\# epochs} & \textbf{learning rate} & \textbf{optimizer} & \textbf{loss} & \textbf{metric} \\ \hline
    Spherical & 64 & 200 & $10^{-1}$ & SGD + momentum & CE loss & Acc. \\
    Darcy Flow & 10 & 200 & $10^{-2}$ & SGD + momentum & Lp loss & Lp loss \\
    NinaPro & 128 & 200 & $10^{-1}$ & SGD + momentum & Focal loss & Acc. \\
    ECG & 1024 & 200 & $10^{-1}$ & SGD + momentum & CE loss & F1  \\
    Satellite & 256 & 200 & $10^{-1}$ & SGD + momentum & CE loss & Acc. \\
    \hline
    \end{tabular}
    \caption{List of hyperparameters and metrics for various NB360 datasets}
    \label{tab:nb360_hyperparams}
\end{table}

\noindent \textbf{Expert models: }We use the same set of hand-crafted models as described in NAS-Bench-360 \cite{tu2021bench}. S2CNN \cite{cohen2018spherical} for Spherical CIFAR-100, Fourier Neural Operator (FNO) Network \cite{li2020fourier} for Darcy flow, Attention-based model \cite{josephs2020semg} for Ninapro DB5, ResNet-1D \cite{hong2020holmes} for ECG and, ROCKET \cite{dempster2020rocket} for the satellite dataset.

\begin{figure}
    \centering
    \includegraphics[width=\textwidth]{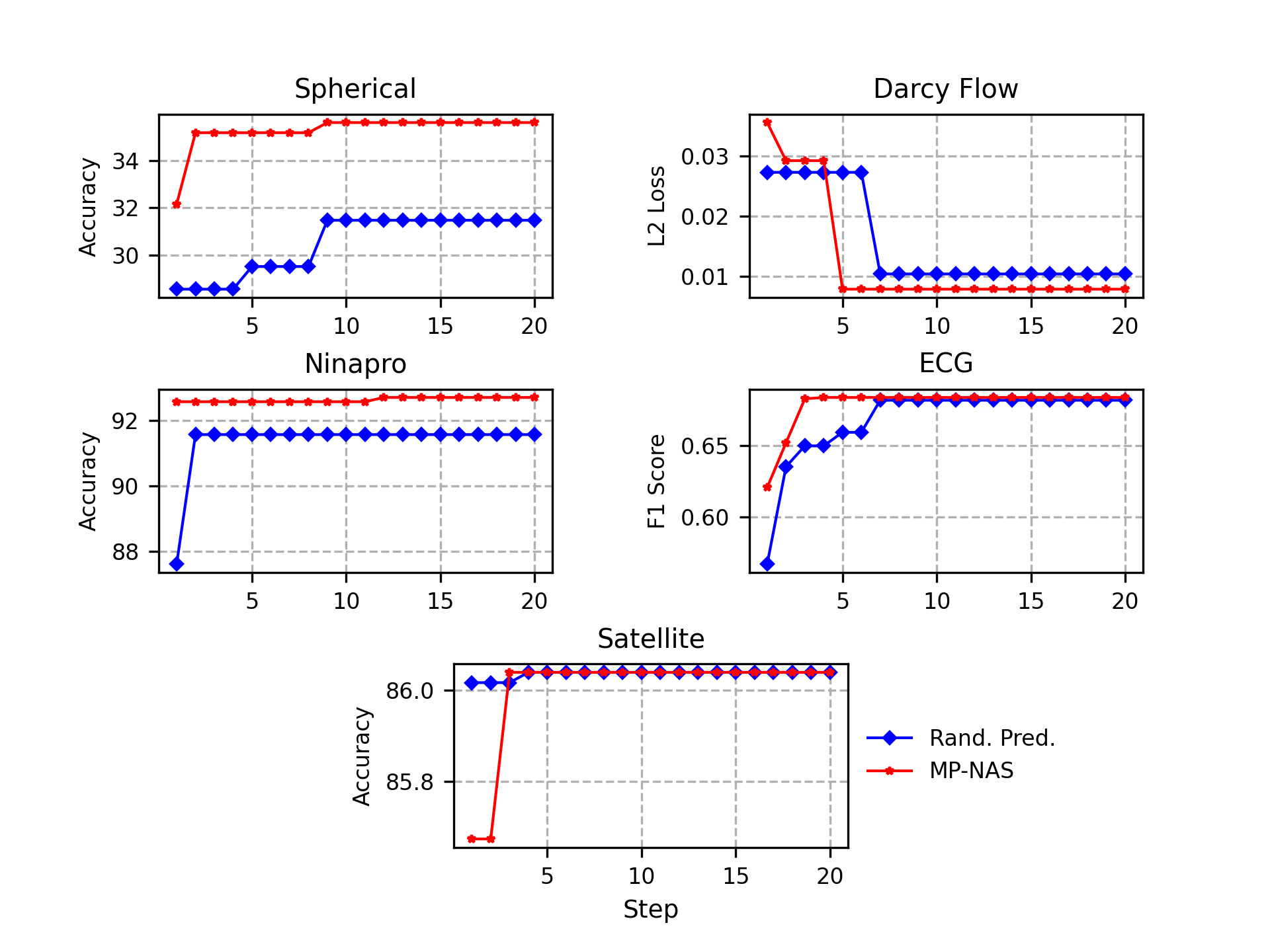}
    \caption{NAS performance comparison of \mlnas with randomly initialized predictor-based NAS.}
    \label{fig:nb360_evolution}
\end{figure}

\noindent \textbf{Comparison between random init. and meta init.: }We compare the NAS performance of our approach with predictor-based NAS where the predictor isn't meta-learnt. We observe not only do we get better accuracy after 20 steps we do so much faster -- usually within first ten steps of the search, whereas random initialized predictor takes longer. The per-dataset NAS evolution can be visualized in Figure~\ref{fig:nb360_evolution}.

\newpage

\section{Additional figures}

\begin{figure}[h!]
    \includegraphics[width=8.975cm]{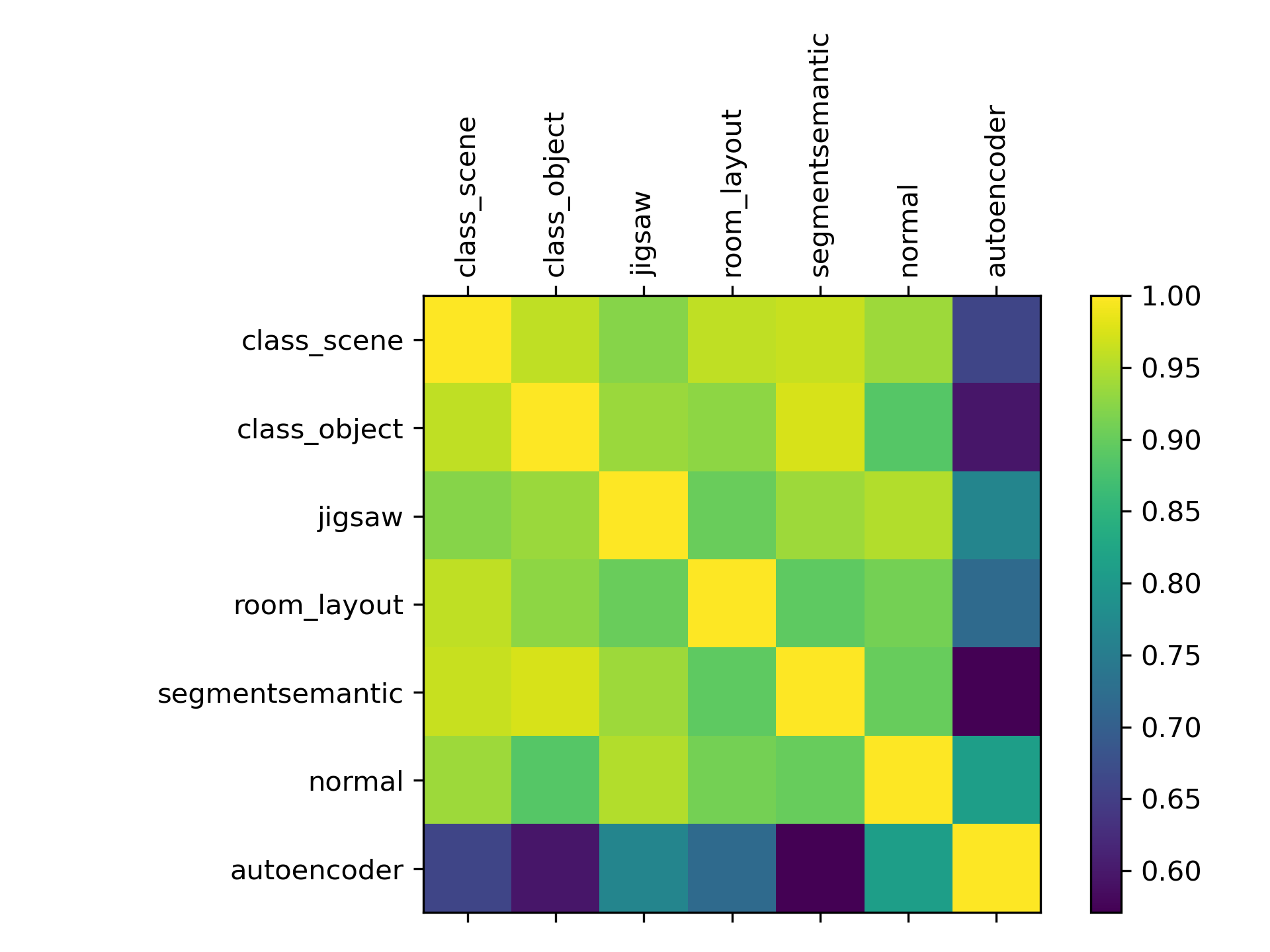}
    \includegraphics[width=8.0cm]{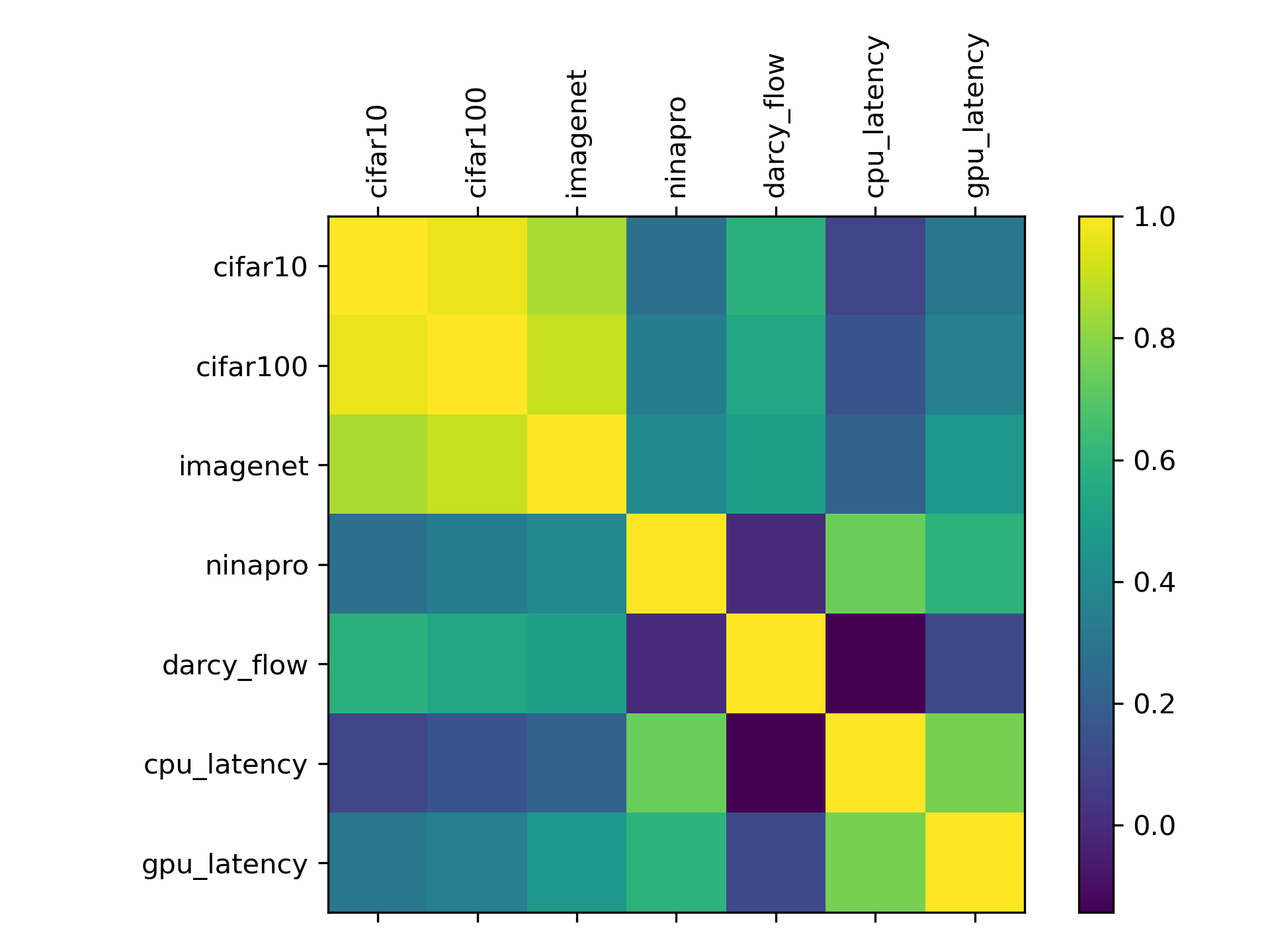}
    \caption{Correlation in accuracies across different datasets in TB101 and NB201.}
    \label{fig:corr}
\end{figure}

We present a detailed correlation matrix for performance of models from NB201 and TB101 on the 7 available tasks for which full NAS data exist (coming from different NAS benchmarks) in Figure~\ref{fig:corr}.

\label{app:d}

\end{document}